# Real-time Scene Segmentation Using a Light Deep Neural Network Architecture for Autonomous Robot Navigation on Construction Sites

Khashayar Asadi. Ph.D. Student, SM.ASCE[1], Pengyu Chen. MSCS Student[2], Kevin Han Ph.D. M.ASCE[3], Tianfu Wu[4], and Edgar Lobaton[4]

[1] Department of Civil, Construction, and Environmental Engineering, North Carolina State University, 2501 Stinson Dr, Raleigh, NC 27606; email: kasadib@ncsu.edu
[2] Department of Computer Science, Columbia University in the City of New York, Mudd Building, 500 W 120th St New York, NY 10027
[3] Department of Civil, Construction, and Environmental Engineering, North Carolina State University, 2501 Stinson Dr, Raleigh, NC 27606
[4] Department of Electrical and Computer Engineering, North Carolina State University, 890 Oval Drive, Raleigh, NC 27606

## ABSTRACT

Camera-equipped unmanned vehicles (UVs) have received a lot of attention in data collection for construction monitoring applications. To develop an autonomous platform, the UV should be able to process multiple modules (e.g., context-awareness, control, localization, and mapping) on an embedded platform. Pixel-wise semantic segmentation provides a UV with the ability to be contextually aware of its surrounding environment. However, in the case of mobile robotic systems with limited computing resources, the large size of the segmentation model and high memory usage requires high computing resources, which a major challenge for mobile UVs (e.g., a small-scale vehicle with limited payload and space). To overcome this challenge, this paper presents a light and efficient deep neural network architecture to run on an embedded platform in real-time. The proposed model segments navigable space on an image sequence (i.e., a video stream), which is essential for an autonomous vehicle that is based on machine vision. The results demonstrate the performance efficiency of the proposed architecture compared to the existing models and suggest possible improvements that could make the model even more efficient, which is necessary for the future development of the autonomous robotics systems.

## INTRODUCTION

In the past decade, the construction industry has struggled to improve its productivity while the manufacturing industry has experienced a dramatic increase (Changal, S., Mohammad, A., and van Nieuwland 2015; Shakeri et al. 2015). The deficiency of advanced automation in construction is one possible reason (Asadi and Han 2018). On the other hand, construction progress monitoring has been recognized as one of the key elements that lead to the success of a construction project (Asadi et al. 2019b). Although there were various attempts by researchers to automate construction progress monitoring (Boroujeni and Han 2017; Bosché et al. 2015; Han and Golparvar-Fard 2017; Kropp et al. 2018; Noghabaei et al. 2019), in the present state, the monitoring task is still performed by site managers through on-site data collection and analysis, which are time-consuming and prone to errors (Balali et al. 2018; Kayhani et al. 2019; Yang et al. 2015). If automated, the time spent on data collection can be better spent by the project management team, responding to any progress deviations by making timely and effective decisions. The use of Unmanned ground and aerial Vehicles (UVs) on construction sites has dramatically grown in the past few years (Asadi et al. 2018b; Ham et al. 2016). This growth can potentially automate the data collection required for visual data analytics that will automate the progress inference process from previous studies (Han et al. 2018).

The authors' previous research on an integrated mobile robotic system (Asadi et al. 2018a) presents an unmanned ground vehicle (UGV) that runs multiple modules and enables future development of an autonomous UGV. In (Asadi et al. 2018a), NVIDIA Jetson TX1 boards (denoted as Jetson boards) (NVIDIA 2017) are used to process simultaneous localization and mapping, motion planning and control, and context awareness (via semantic image segmentation) modules of the system. The two major bottlenecks of this platform, in terms of computational loads, were SLAM and segmentation. For this reason, there was a designated Jetson board for each of these tasks. This is a major challenge in developing an autonomous robot because it increases the size and weight, especially with added batteries. Moreover, the problem becomes even more challenging if this robotics system was to be applied to an unmanned aerial vehicle (UAV).

This previous work implements ENet (Paszke et al. 2016) as the semantic segmentation method. The ENet model was designed to run on embedded boards, which makes this model more applicable to mobile robotics systems. However, the segmentation task by this model had the heaviest computational load, which made the authors restrict the speed of the UGV for real-time performance necessity. Combining the SLAM and Context-Awareness Modules into the same Jetson TX1 would help to mitigate this problem, but the large size of the segmentation model and the high memory usage, make this solution practically unfeasible. The major goal of this paper is to propose a deep convolution neural network (CNN) which reduces the computational load (model size) in order to reduce the latency by running multiple modules on the same Jetson while maintaining the accuracy.

**METHOD**

Real-time semantic segmentation algorithms are used to understand the content of images and find target objects in real-time which are crucial tasks in mobile robotic systems. In this part, a new convolutional neural network model architecture, as well as the strategies and techniques for its training, compressing and data processing will be introduced. Adam Optimizer (Kingma and Ba 2014) is used to train the model and cross entropy is utilized as loss function. Also, a new pixel-level annotated dataset has been generated and validated for real-time and mobile semantic segmentation in a construction engineering environment. Figure 1 shows an overview of the components of the proposed algorithm. In the following, first, factorized convolution block as a core block is described which the proposed model is built on. We then illustrate the network structure followed by a model compressing method description.

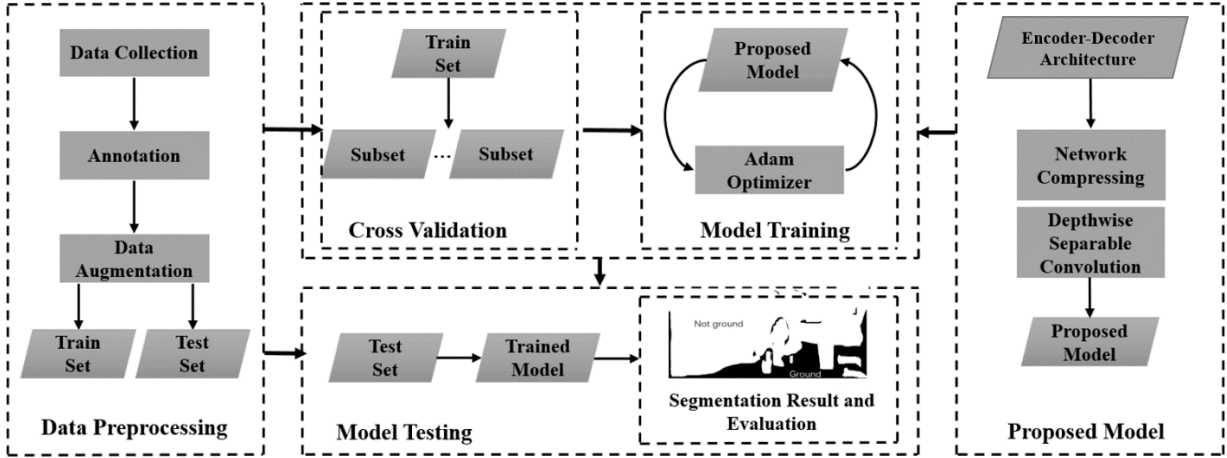

**Figure 1. An overview of the proposed model for real-time navigable space segmentation.**

**Factorized Convolution Block**

To optimize the amount of parameters in a CNN model while maintaining its performance, cut down the model sizes, and run the model faster, Depthwise Separable Convolution from MobileNet (Howard et al. 2017) is applied. Depthwise Separable Convolution is a form of factorized convolutions which factorizes a standard convolution into two separate operations - a depthwise convolution and a point-wise convolution. While the normal convolution operates over every channel of the input feature map, Depthwise Separable Convolution applies a single filter on every channel and use a point-wise convolution, ($1\times1$) kernel convolution, and linearly combines the outputs. For a standard convolution layer whose kernel size is $K$, it takes a feature map of size $H_{in} \times W_{in} \times C_{in}$ as input where $H_{in}$, $W_{in}$, $C_{in}$ are height, width, and channels respectively, and outputs a feature map of size $H_{out} \times W_{out} \times C_{out}$. The computational cost is computed using Equation 1.

$$K \times K \times C_{in} \times C_{out} \times H_{in} \times W_{in} \quad (1)$$

Equation 2, shows the computational cost for separable depthwise convolution, considering the same feature map.

$$K \times K \times C_{in} \times H_{in} \times W_{in} + C_{in} \times C_{out} \times H_{in} \times W_{in} \quad (2)$$

By transforming the original convolution operation into depthwise separable convolution, the reduction in computation cost is calculated by Equation 3.

$$\frac{K \times K \times C_{in} \times H_{in} \times W_{in} + C_{in} \times C_{out} \times H_{in} \times W_{in}}{K \times K \times C_{in} \times C_{out} \times H_{in} \times W_{in}} \quad (3)$$

In the proposed model, depthwise separable convolutional layers replace the standard convolution layers in a CNN model. This replacement reduces the computational cost. For instance, a convolutional layer with $3\times3$ filters is substituted with depthwise separable convolutional layer. The reduced computational cost is *0.13*, which means that the model uses almost 8 times less computation than standard computations (see Equation 4). This reduction in computation results in a negligible reduction in the model's accuracy (see EXPERIMENTAL SETUP AND RESULTS Section).

$$\frac{1}{C_{out}} + \frac{1}{K^2} = \frac{1}{64} + \frac{1}{9} \tag{4}$$

In the proposed model, depthwise separable convolutional layers replace the standard convolution layers in a CNN model. This replacement reduces the computational cost. By utilizing the strength of depthwise separable convolution, we proposed a new residual block based on the traditional residual block. The block has two branches. One branch consists of two *1×1* convolutional layers (the first one is for projection and reducing the dimensionality which is located before the depthwise separable convolution layer and the second one is placed afterward which is for expansion), a depthwise convolutional layer, and a batch normalization layer at the end. The other one is a shortcut branch which outputs an identical feature map as its input. If the type of the block is Downsample, a MaxPooling layer is added to the shortcut branch. At the end of the block, the outputs from the two branches are element-wise added. The structure of the block is shown in Figure 2.

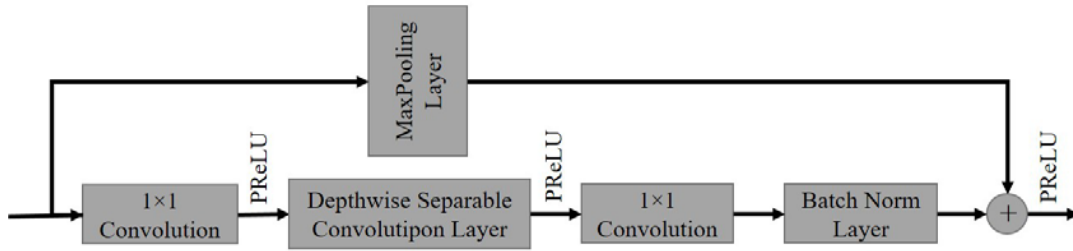

**Figure 2. Factorized Convolutional Block**

**Architecture Design.**
The proposed model has a light model that has been optimized for a mobile device and real-time semantic segmentation applications. The model has an encoder-decoder architecture, which consists of several Factorized Convolution Blocks. The architecture of our network is presented in Table 1. In the first Initial block where we followed the ENet architecture, we concatenated the output of a *3×3* convolutional layer with the stride of two and a MaxPooling layer to decrease the feature size at an early stage. The Downsample and Standard blocks follow the structure we illustrated in Factorized Convolution Block part. As for the Upsample and LastConv blocks, the same configurations are applied as ENet.

**Model Compressing**
A well-trained CNN model usually has significant redundancy among different filters and feature channels. Therefore, an efficient way to cut down both computational cost and model size is to compress the model (Li et al. 2016). L1 norm of the filter is used to evaluate the efficiency of the filter. The L1 norm of a vector x is calculated using Equation 5, where $x_i$ represents each value of the kernel.

$$||x||_1 = \sum |x_i| \tag{5}$$

If the L1 norm is close to zero, it generates negligible output because the output feature map values tend to be close to zero. To improve efficiency, removing these filters would reduce the computational expense and model size. Network compressing consists of the following steps: For each filter, the L1 value is calculated and filters are sorted by L1 values. Then, the filters and corresponding feature maps are removed and the whole network is fine-tuned. However, it is hard to predict how many filters and which filters can be removed without damaging the performance. By calculating and sorting the L1 values of every filter in the model, the 128 filters (denoted as 128-kernels) generally, have smaller $L^1$ values and have the largest number of redundant filters. 128-kernels are less sensitive to the changes compared to other kernels. Therefore, removing 128- kernels have a negligible impact on the performance. As a result, these filters could be removed without harming the accuracy. As it is shown in Table 1 half of the filters of 128-kernels are removed (is indicated with * in the table) and the model is fine-tuned on the same dataset.

**Table 1. Model Architecture. Output sizes are given for an example RGB input of 512×256.**

| Block | Type | Input Size | Output Size |
|---|---|---|---|
| 1 | Initial | 512 × 256 × 3 | 256 × 128 × 16 |
| 2 | Downsample | 256 × 128 × 16 | 128 × 64 × 64 |
| 3 - 6 | Standard | 128 × 64 × 64 | 128 × 64 × 64 |
| 7 | Downsample | 128 × 64 × 64 | 64 × 32 × 128(64*) |
| 8 - 24 | Standard | 64 × 32 × 128(64*) | 64 × 32 × 128(64*) |
| 25 | Upsample | 64 × 32 × 128(64*) | 128 × 64 × 64 |
| 26 | Standard | 128 × 64 × 64 | 128 × 64 × 64 |
| 27 | Standard | 128 × 64 × 64 | 128 × 64 × 64 |
| 28 | Upsample | 128 × 64 × 64 | 256 × 128 × 16 |
| 29 | Standard | 256 × 128 × 16 | 256 × 128 × 16 |
| 30 | full Conv | 256 × 128 × 16 | 512 × 256 × 2 |

**EXPERIMENTAL SETUP AND RESULTS**

**Dataset Construction and Training Strategy**
A new dataset has been constructed for the proposed model. The main objective of our classification task is to segment all objects from navigable spaces in the scene, which is why the proposed model has only two classes. To segment objects even in a new environment, the data is collected from three completely different environments including construction sites, parking lots, and roads (1000 images). The Cityscapes public dataset (Cordts et al. 2016) is also used for training and testing process. This dataset can prevent over-fitting of the proposed model. Therefore, it was used for training and cross-validation. Cityscape has 5,000 frames that have high-quality pixel-level annotations. As it has 30 classes, these labels are grouped into 2 classes for our task (ground and not ground)

Tensorflow (Google 2015) and Keras framework are used to implement the algorithm. CUDA (Nickolls et al. 2008) and CuDNN (Chetlur et al. 2014) are also utilized for accelerating the computations. The whole training process took about 12 hours. Since Asadi et al. (Asadi et al. 2018a) have already validated ENet implementation on the embedded platform in real-time, this paper focuses on comparing the performance of ENet and the proposed model on a server with the following specification: 128 GB RAM, Intel Xeon E5 processor, and two GPUs - NVIDIA

Tesla K40c and Telsa K20c. Figure 3 shows sample images (top) and their corresponding labels (bottom).

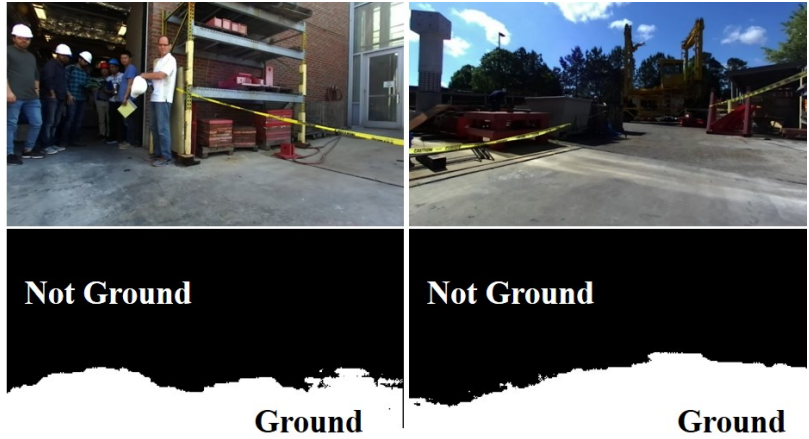

**Figure 3. Examples of proposed dataset image (top), corresponding labels (bottom).**

**Test Results**
For testing and evaluation, 150 images are used from three scenes, including a parking lot, road, and construction site. Due to the differences between the three scenes, models performance is varied and highly related to the complexity of images and categories of objects. The test results are shown in Table 2.

**Table 2. Proposed vs ENet Accuracy in Test Data**

|  | ENet | Proposed Model |
|---|---|---|
| Recall | 0.9890 | 0.9908 |
| Precision | 0.8855 | 0.8384 |
| Accuracy | 0.9435 | 0.9231 |

**Model Comparison**
The results show that only 2% percent of accuracy is lost, on the other hand, the complexity of the network is reduced greatly by decreasing the parameters and the feature layers by half. The model size drops from 2225 KB to 1068 KB. The comparison between the two models is shown in Table 3. The inference time is directly related to computation cost in the network. The proposed model's inference time decreased by 18%, although the number of parameters decreased significantly (see Table 3. There are two tasks performed by the GPU: creating the kernel and evaluating the model. The proposed model only decreases the evaluation time which is why the inference time does not decrease at the same rate. In other words, the inference time depends greatly on other computation factors rather than just the evaluation of the neural network model (i.e., the cost of launching the GPU kernel starts to dominate the computation time).

    The average inference time per frame has reduced from 44.7 ms to 36.5 ms per frame. This reduction in the inference time means that the maximum input frame rate for real-time performance can be increased by 5 fps (from 22 fps in ENet to 27 fps in proposed model). Although the proposed segmentation model has lower inference time compare to the ENet

method, the main contribution of the proposed model is reducing the model size (more than 50%). This reduction enables multiple modules to be run on the same Jetson TX1, which will reduce the latency caused by integrating multiple Jetson TX1s through the wired network.

**Table 3. Proposed Model vs ENet**

|  | Proposed Model | ENet |
|---|---|---|
| Average inference time per frame (ms) | 36.5 | 44.7 |
| Total parameters | 227393 | 371116 |
| Max input frame rate for real-time performance | 27.4 | 22.4 |
| Model size (KB) | 1068 | 2225 |

**CONCLUSION**

This paper presents an efficient semantic segmentation model that can be run in real-time on multiple embedded platforms that are integrated as a system for navigable space segmentation. The main contributions of this paper are 1) a new pixel-level annotated dataset for real-time and mobile semantic segmentation in construction environments and combining with transfer learning to deal with the limited number of training dataset and 2) an efficient semantic segmentation method with a smaller model size and faster inference speed for future development of autonomous robots on construction sites. Although the focus of this study is on reducing the model size to enable running multiple modules on the same Jetson TX1, the inference time is also decreased, which increases the frame rate of the segmentation process. 50% reduction in the model size is a significant contribution, which enables multiple modules to be combined and run on the same Jetson TX1. By doing this, the latency caused by integrating multiple Jetson TX1 through the wired network will reduce drastically (Asadi et al. 2019a).